\documentclass[11pt]{article}

\usepackage[margin=1in]{geometry}
\usepackage[T1]{fontenc}
\usepackage[utf8]{inputenc}
\usepackage{lmodern}
\usepackage{microtype}

\usepackage{amsmath,amssymb}
\usepackage{booktabs}

\usepackage{graphicx}
\usepackage{float}

\usepackage{algorithm}
\usepackage{algorithmic}
\usepackage{listings}
\usepackage{newfloat}

\usepackage[round,sort&compress]{natbib}

\usepackage[hidelinks]{hyperref}
\usepackage{url}

\floatstyle{ruled}
\newfloat{listing}{tb}{lst}
\floatname{listing}{Listing}

\title{LLM-Guided Evolutionary Program Synthesis for Quasi-Monte Carlo Design}

\author{Amir Sadikov}
\date{}

\begin{document}

\maketitle

\begin{abstract}
Low-discrepancy point sets and digital sequences underpin quasi-Monte Carlo (QMC) methods for high-dimensional integration. We cast two long-standing QMC design problems as program synthesis and solve them with an LLM-guided evolutionary loop that mutates and selects code under task-specific fitness: (i) constructing finite 2D/3D point sets with low star discrepancy, and (ii) choosing Sobol’ direction numbers that minimize randomized QMC error on downstream integrands. Our two-phase procedure combines constructive code proposals with iterative numerical refinement. On finite sets, we rediscover known optima in small 2D cases and set new best-known 2D benchmarks for $N \ge 40$, while matching most known 3D optima up to the proven frontier ($N \le 8$) and reporting improved 3D benchmarks beyond. On digital sequences, evolving Sobol’ parameters yields consistent reductions in randomized quasi-Monte Carlo (rQMC) mean-squared error for several 32-dimensional option-pricing tasks relative to widely used Joe--Kuo parameters, while preserving extensibility to any sample size and compatibility with standard randomizations. Taken together, the results demonstrate that LLM-driven evolutionary program synthesis can automate the discovery of high-quality QMC constructions, recovering classical designs where they are optimal and improving them where finite-N structure matters. Data and code are available at \url{https://github.com/hockeyguy123/openevolve-star-discrepancy.git}.
\end{abstract}

\section{Introduction}
Numerical integration in high dimensions is a cornerstone of modern science and engineering. While standard Monte Carlo (MC) methods offer a robust approach, their convergence rate, governed by the Central Limit Theorem, is often insufficient for applications requiring high precision \cite{Glasserman2003}. Quasi-Monte Carlo (QMC) methods provide a compelling alternative by replacing pseudorandom samples with deterministic, highly uniform point sets \cite{Dick2010, Owen1995}. The uniformity of these sets is quantified by their discrepancy, with lower values corresponding to more evenly distributed points. The Koksma--Hlawka inequality provides the theoretical underpinning for QMC, guaranteeing that the integration error is bounded by the product of the variation of the integrand and the star discrepancy of the point set \cite{Koksma1964, Hlawka1961}.

This relationship has fueled decades of research into the discovery and construction of low-discrepancy sets. A primary challenge in this field is the ``\emph{ab initio}'' construction of a finite point set of $N$ points in $d$ dimensions that minimizes star discrepancy. This is a problem of combinatorial complexity, and while optimal solutions have been found in 2D for $N \le 21$ and in 3D for $N \le 8$ \cite{clement2024}, the problem remains largely open for larger $N$ and higher dimensions.

A related challenge is the construction of infinite low-discrepancy sequences, such as those by Halton, Hammersley, and Sobol'. Among these, Sobol' sequences are particularly prominent due to their excellent distribution properties and efficient generation \cite{Sobol1967}. However, their quality is highly dependent on a set of integer parameters known as direction numbers, and finding optimal parameters that ensure uniformity across all low-dimensional projections is a difficult combinatorial search problem \cite{Joe2008}.

Recent breakthroughs in Large Language Models (LLMs) have demonstrated their remarkable capabilities in code generation, logical reasoning, and pattern recognition \cite{Gemini2025}. This has motivated the development of automated scientific discovery systems, such as AlphaEvolve, which leverage LLMs to navigate complex search spaces \cite{Novikov2025}. In this work, we utilize the OpenEvolve framework \cite{openevolve}, an open-source implementation based on the principles of AlphaEvolve, to tackle the aforementioned challenges in discrepancy theory. We treat the construction of these mathematical objects as a program synthesis problem within an evolutionary framework \cite{Koza1994}. The LLM acts as an intelligent mutation operator, iteratively modifying code that generates candidate solutions based on feedback from a fitness function.

This paper presents the successful application of this LLM-driven evolutionary approach to two fundamental problems:
\begin{enumerate}
    \item Discovering finite 2D and 3D point sets with state-of-the-art low star discrepancy.
    \item Searching for superior Sobol' direction numbers to minimize randomized quasi-Monte Carlo (rQMC) integration error in the high-dimensional context of pricing exotic financial options \cite{Paskov1995}.
\end{enumerate}
Our results show that this methodology can match and, in several important cases, surpass the best known human-derived solutions. This suggests that LLM-driven evolutionary search is a promising new paradigm for exploration and discovery in computational mathematics.

\section{Theoretical Background}
\subsection{Star Discrepancy}
Star discrepancy is the most common measure for quantifying the uniformity of a point set within the $d$-dimensional unit hypercube, $[0,1]^d$ \cite{Owen1995}. It captures the largest deviation between the volume of an axis-aligned ``anchor box'' and the fraction of points contained within it.

\paragraph{Definition 1 (Star Discrepancy).} Let $P = \{\mathbf{x}_1, \dots, \mathbf{x}_N\}$ be a set of $N$ points in $[0,1]^d$. An anchor box $[\mathbf{0}, \mathbf{q})$ for any $\mathbf{q} = (q_1, \dots, q_d) \in [0,1]^d$ is the hyperrectangle $[0, q_1) \times \dots \times [0, q_d)$. The star discrepancy $D_N^*$ of the set $P$ is defined as:
\begin{equation}
D_N^*(P) = \sup_{\mathbf{q} \in [0,1]^d} \left| \frac{\#(P \cap [\mathbf{0}, \mathbf{q}))}{N} - \text{Vol}([\mathbf{0}, \mathbf{q})) \right|
\label{eq:star_discrepancy_abbr}
\end{equation}
Here, the supremum is taken over all possible anchor boxes. A small $D_N^*$ value implies that for any anchor box, the fraction of points falling within it is a good approximation of its volume, indicating high uniformity. The practical importance of star discrepancy is cemented by the Koksma--Hlawka inequality \cite{Koksma1964, Hlawka1961}, which bounds the error of QMC integration:
\begin{equation}
\left| \int_{[0,1]^d} f(\mathbf{u})\, d\mathbf{u} - \frac{1}{N} \sum_{i=1}^{N} f(\mathbf{x}_i) \right| \le V(f) \cdot D_N^*(P)
\label{eq:koksma_hlawka}
\end{equation}
where $V(f)$ is the total variation of the function $f$ in the sense of Hardy and Krause. This inequality guarantees that point sets with lower star discrepancy lead to smaller integration error bounds.

\subsection{Sobol' Sequences and Direction Numbers}
Sobol' sequences are a class of low-discrepancy sequences that are particularly effective for QMC integration \cite{Sobol1967, Dick2010}. They are constructed using the properties of primitive polynomials over the finite field of two elements, $\mathbb{F}_2$.

\paragraph{Definition 2 (Sobol' Sequence Construction).} For each dimension $j \ge 1$, a primitive polynomial over $\mathbb{F}_2$ of degree $s_j$ is chosen \cite{Dick2010}:
\begin{equation}
P_j(z) = z^{s_j} + a_{1,j}z^{s_j-1} + \dots + a_{s_j-1,j}z + 1
\label{eq:primitive_poly}
\end{equation}
where the coefficients $a_{k,j}$ are either 0 or 1. From this polynomial, a sequence of positive, odd integers called direction numbers $m_{k,j}$ (for $k = 1, \dots, s_j$) are chosen freely. 
% Subsequent direction numbers ($k > s_j$) are generated via the recurrence relation:
% \begin{equation}
% m_{k,j} = 2a_{1,j}m_{k-1,j} \oplus \dots \oplus 2^{s_j}m_{k-s_j,j} \oplus m_{k-s_j,j}
% \label{eq:direction_numbers_recurrence}
% \end{equation}

% For $k>s_j$, subsequent direction numbers (expressed as fixed-point binary fractions $v_{k,j}\!=\!m_{k,j}/2^k$) are generated by the canonical Sobol' recurrence (e.g., \cite{Joe2008,Dick2010}):
% \begin{equation}
% v_{k,j} \;=\; a_{1,j}v_{k-1,j}\;\oplus\; \cdots \;\oplus\; a_{s_j-1,j}v_{k-s_j+1,j}\;\oplus\; v_{k-s_j,j}\;\oplus\; \big(v_{k-s_j,j} \gg s_j\big)
% \label{eq:direction_numbers_recurrence}
% \end{equation}
% where $\oplus$ denotes bitwise XOR on the binary fixed-point representation and $\gg s_j$ is a right bit-shift by $s_j$ places. For $j\!=\!1$, we set $v_{k,1}=2^{-k}$ (van der Corput base-2).
For $k>s_j$, subsequent direction numbers (expressed as fixed-point binary fractions $v_{k,j}=m_{k,j}/2^k$) are generated by the canonical Sobol' recurrence (e.g., \cite{Joe2008,Dick2010}):
\begin{equation}
v_{k,j} \;=\; a_{1,j}v_{k-1,j}\;\oplus\; a_{2,j}v_{k-2,j}\;\oplus\;\cdots\;\oplus\; a_{s_j-1,j}v_{k-s_j+1,j}\;\oplus\;\big(v_{k-s_j,j}\gg s_j\big)
\label{eq:direction_numbers_recurrence}
\end{equation}
where $\oplus$ denotes bitwise XOR on the binary fixed-point representation and $\gg s_j$ is a right bit-shift by $s_j$ places. For $j=1$, we set $v_{k,1}=2^{-k}$ (van der Corput base-2).

The $j$-th coordinate of the $i$-th point in the sequence, $x_{i,j}$, is then generated using Gray-code bits:
\begin{equation}
x_{i,j} \;=\; \bigoplus_{k\ge 1} g_k\, v_{k,j},
\qquad
\text{with } i=\sum_{k\ge 1} i_k 2^{k-1},\;\;
g \;=\; i \oplus (i \gg 1) \;=\; \sum_{k\ge 1} g_k 2^{k-1}.
\label{eq:sobol_point_gen}
\end{equation}
The quality of the Sobol' sequence, particularly the uniformity of its low-dimensional projections, is critically dependent on the choice of the primitive polynomials and the initial direction numbers ($m_1, \dots, m_s$). The work of \citet{Joe2008} provides a widely used set of these parameters that serve as a strong baseline. In tables we encode the polynomial coefficients $a_{k,j}$ as an integer $A_j$ (binary bitmask).
 
\section{Related Work}
Classical approaches for generating low-discrepancy sets are primarily number-theoretic. Foundational methods include Halton, Hammersley, and Sobol' sequences, which are designed to achieve superior asymptotic uniformity compared to random sampling. While powerful, these classical constructions are not always optimal for a finite number of points $N$. Mathematical programming has been used to find provably optimal sets, though these approaches are computationally intensive and limited to small instances \cite{clement2024}. Heuristic methods, such as genetic algorithms and threshold accepting, have been applied to tackle larger instances by searching the space of point configurations \cite{clement2023}.

More recently, machine learning techniques have been introduced to this domain. Message-Passing Monte Carlo (MPMC) leverages Graph Neural Networks (GNNs) to transform random initial points into low-discrepancy configurations, achieving state-of-the-art results by directly optimizing point coordinates \cite{Rusch2024}. Our work differs by framing the task as a program synthesis problem rather than direct coordinate optimization.

A related line of research focuses on optimizing the parameters of Sobol' sequences. The quality of a Sobol' sequence is critically dependent on a set of initialization parameters known as direction numbers. The direction numbers published by \citet{Joe2008} are a widely-used standard, derived from an extensive computational search to find parameters that ensure high uniformity in two-dimensional projections by minimizing a quality measure known as the $t$-value. Subsequent work has focused on further improving these parameters or guaranteeing quality for specific projection properties crucial for applications such as computer graphics \cite{Bonneel2025}.

Our approach is most closely related to the emerging paradigm of using Large Language Models (LLMs) for automated scientific discovery. \citet{Novikov2025} introduces AlphaEvolve, a framework that combines LLMs with an evolutionary search, treating algorithm discovery as a program evolution problem where the LLM functions as an intelligent mutation operator and receives feedback from a fitness function. This method has successfully discovered novel algorithms for fundamental problems, from matrix multiplication \cite{Fawzi2022} to open mathematical conjectures \cite{RomeraParedes2024}. Our work is directly inspired by these principles, applying a similar evolutionary loop to the specific mathematical challenges of discovering low-discrepancy sets and optimizing Sobol' sequences.

\section{Methodology}
Our approach is based on the OpenEvolve framework, an open-source implementation of the principles demonstrated by AlphaEvolve \cite{Novikov2025}. It frames the search for novel mathematical constructs as an evolutionary search over a population of programs that generate them \cite{Koza1994}. The LLM serves as a sophisticated mutation operator, guided by a fitness function.

The evolutionary loop \cite{RomeraParedes2024} proceeds as follows:
\begin{enumerate}
    \item \textbf{Initialization:} The process begins with a population of ``parent'' programs. These programs are code snippets in Python that generate a candidate solution. The initial population can be seeded with simple heuristics or well-known constructions.
    \item \textbf{Evaluation:} Each program in the population is executed, and its output is evaluated by a fitness function. The fitness function returns a scalar score quantifying the quality of the solution (e.g., lower rQMC MSE or star discrepancy is better).
    \item \textbf{Selection and Prompting:} High-performing programs are selected to serve as parents. A detailed prompt is then constructed for the LLM, including the parent program's source code, its fitness score, and an instruction tasking the LLM with generating a variation that will improve upon the score. The prompt also includes code from other high-performing ``inspirations'' to encourage crossover as well as guidance from the user (Appendix B).
    \item \textbf{Generation (Mutation):} The LLM receives the prompt and generates a new, modified program. This is the core ``mutation'' step. The LLM's ability to understand code syntax and semantics allows for complex and intelligent modifications.
    \item \textbf{Loop:} The newly generated program is evaluated, its fitness is scored, and it is added to the population. The process then repeats, iteratively refining the population toward better solutions.
\end{enumerate}

\subsection{Experimental Setup}
We used an LLM-guided evolutionary search over a population of Python programs (population 60, multi-island evolution with occasional migration). Parents were chosen by fitness with occasional archive sampling; the LLM rewrote core functions each generation. We ran a fixed compute budget and did not tune hyperparameters; full settings are available in Appendix B.

\subsection{Discovery of Low-Discrepancy Point Sets}
A two-phase strategy was employed to balance broad exploration with fine-tuned optimization.
\begin{itemize}
    \item \textbf{Phase 1: Direct Construction:} The LLM was prompted to generate Python code that directly constructs an $N$-point set in a $d$-dimensional space. The initial parent program in 2D implemented a simple shifted Fibonacci lattice (Appendix B, Listing 1) and in 3D implemented a scrambled Sobol' sequence (Appendix B, Listing 2). This phase encouraged the LLM to explore a wide range of constructive heuristics.
    \item \textbf{Phase 2: Iterative Optimization:} After a sufficient number of iterations, the LLM was prompted to generate Python code that uses iterative optimization routines (e.g., \texttt{scipy.optimize.minimize}) to refine an initial guess. This shifted the search from finding explicit constructions to a direct optimization of the point coordinates.
    \item \textbf{Fitness Function:} The fitness score was $\frac{1}{1+D_N^*}$ where $D_N^*$ is the star discrepancy of the generated point set.
\end{itemize}

\subsection{Star Discrepancy Evaluation}
In 2D we compute $D_N^*$ exactly by scanning extremal anchor boxes on the coordinate-induced grid (overall $O(N\log N)$). In 3D we approximate the supremum with threshold-accepting over randomized axis-aligned boxes. We validated by reproducing published 2D optima for $N\!\le\!20$ and 3D optima for $N\!\le\!8$ to at least $10^{-6}$. Implementation details and checks are in Appendix~A.

\subsection{Discovery of Sobol' Direction Numbers}
\begin{itemize}
    \item \textbf{Program Representation:} The evolved programs are Python functions that return a list of dictionaries. Each dictionary contains the Sobol' parameters (\texttt{s}, \texttt{a}, \texttt{m\_i}) for a single dimension.
    \item \textbf{Initialization:} The initial population contains the following implementation of the direction numbers \cite{Joe2008}.
    \item \textbf{Fitness Function:} The primary fitness metric is $\frac{1}{1 + \mathrm{MSE}}$ where MSE is the mean squared error of an rQMC estimate for a 32-dimensional Asian option price. The MSE is calculated for $N=8192$ points and averaged over 1000 consistent randomizations (left matrix scramble followed by a Cranley--Patterson random shift, i.e., LMS+shift) to ensure robustness and reproducibility. All rQMC comparisons are paired by using identical randomization seeds per method and $N$. The diffusion paths are constructed from $[0,1]^d$ via standard time discretization of geometric Brownian motion with equal timesteps; we report this mapping to clarify effective-dimension effects.
\end{itemize}

\section{Experiments and Results}
\begin{figure*}[t]
\centering
\includegraphics[width=0.95\textwidth]{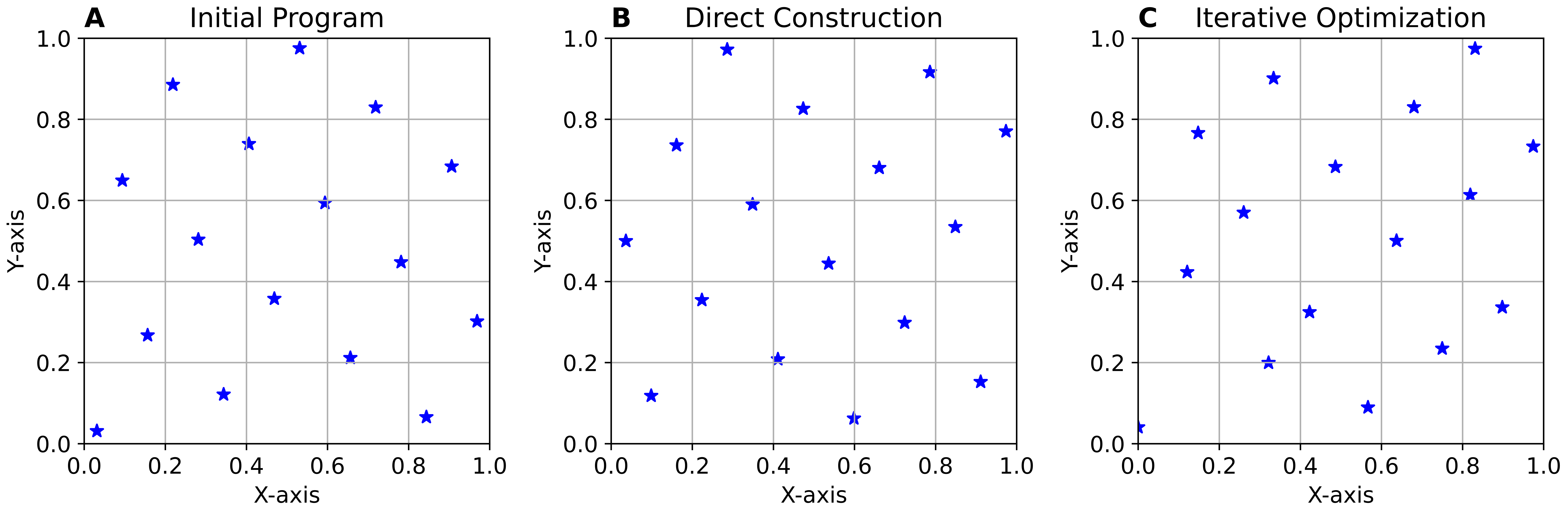}
\caption{Visualization of $N=16$ point set generation in two dimensions. (A) Initial shifted Fibonacci lattice (Discrepancy: 0.0962). (B) Best direct construction found in Phase 1 (Discrepancy: 0.0924). (C) Final optimized point set from Phase 2 (Discrepancy: 0.0744), which is within 0.68\% of the known optimal value of 0.0739.}
\label{fig:discovery_process}
\end{figure*}

\begin{figure*}[t]
\centering
\includegraphics[width=0.95\textwidth]{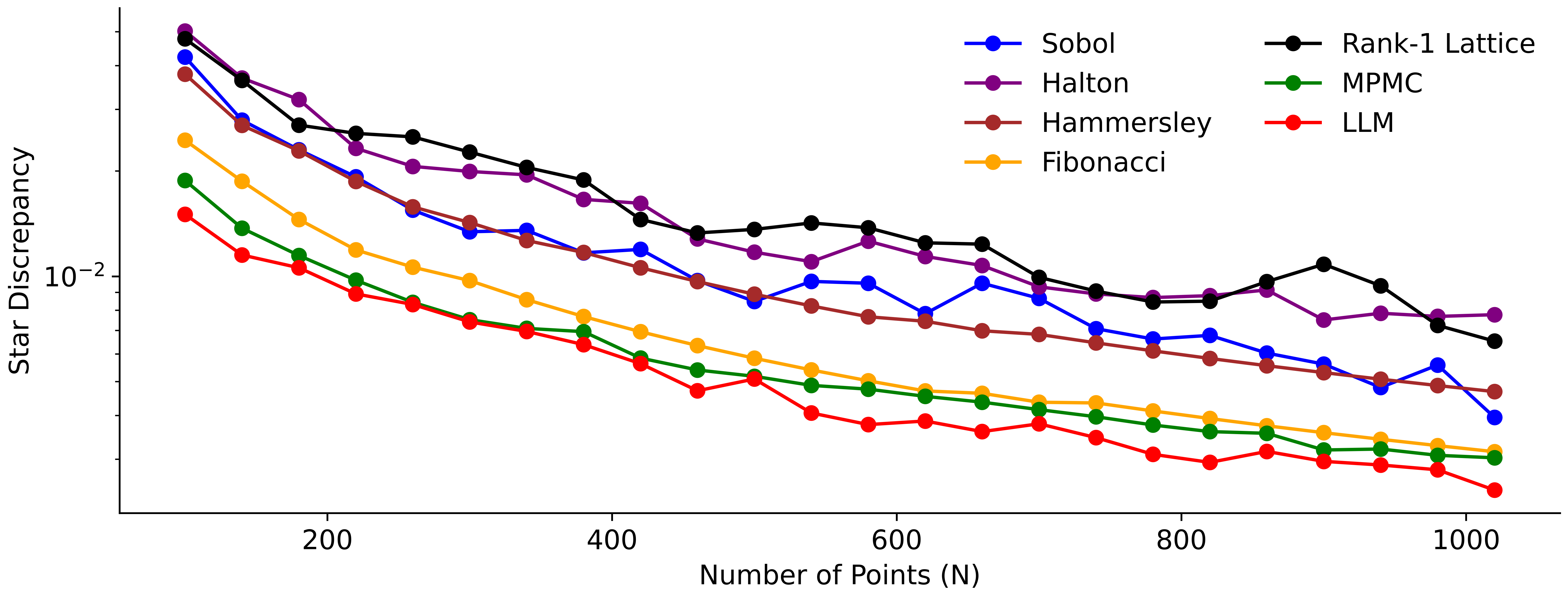}
\caption{The Star Discrepancy $D_N^*$ of Sobol', Halton, Hammersley, Fibonacci, Rank-1-Lattice, MPMC (message passing Monte Carlo), and LLM-evolved sets for increasing number of points $N = 100 \dots 1020$ in 2D.}
\label{fig:n100_point_set}
\end{figure*}

\begin{figure}[t]
\centering
\includegraphics[width=0.5\linewidth]{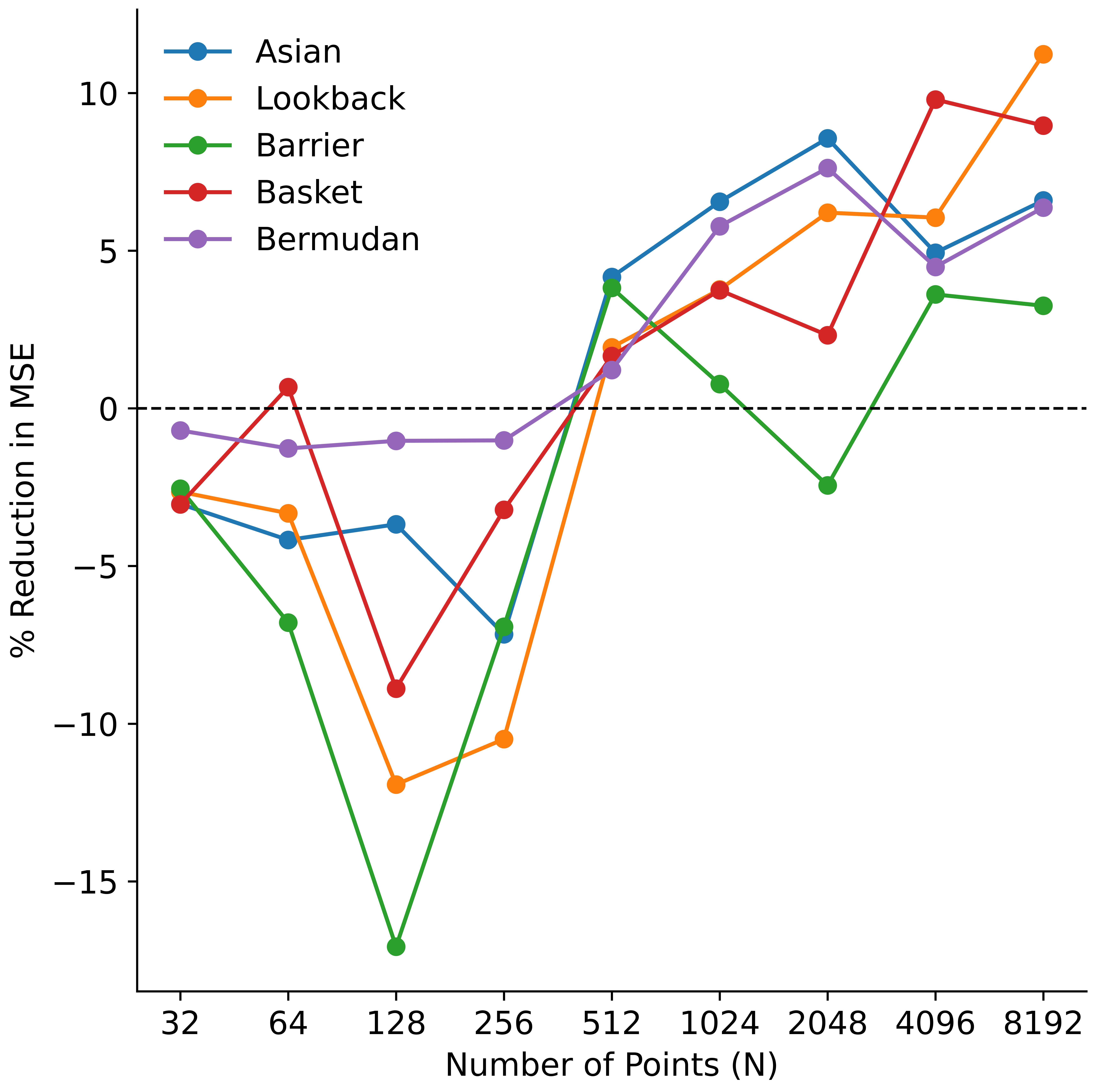}
\caption{The \% reduction in MSE (rQMC integration over 10000 random scrambles and shifts) using Sobol' direction numbers found via LLM evolutionary search vs.\ those of \citet{Joe2008}. The \% reduction in MSE are averaged across all scenarios of that particular option (Appendix~C).}
\label{fig:perf-option}
\end{figure}

\begin{listing}[tb]
\caption{Directly Constructed 16 Point Set ($N=16$)}
\label{lst:direct_con_2d}
\small
\begin{lstlisting}[language=Python]
def construct_star():
    A = np.zeros((N, 2))
    phi=(math.sqrt(5)-1)/2
    for i in range(N):
        A[i, 0]=(i+(1/math.sqrt(3)))/N
        A[i, 1]=((0.5+(i*phi)%1))%1
    return A
\end{lstlisting}
\end{listing}

\begin{table}[h]
\centering
\small
\begin{tabular}{@{}lcccc@{}}
\toprule
\textbf{N} & \textbf{Fibonacci} & \textbf{MPMC} & \textbf{LLM} & \textbf{Cl\'ement et al.} \\ \midrule
1          & 1.0000             & \textbf{0.6180} & \textbf{0.6180}     & \textbf{0.6180}  \\
2          & 0.6909             & \textbf{0.3660} & \textbf{0.3660}     & \textbf{0.3660}  \\
3          & 0.5880             & \textbf{0.2847} & \textbf{0.2847}     & \textbf{0.2847}  \\
4          & 0.4910             & \textbf{0.2500} & \textbf{0.2500}     & \textbf{0.2500}  \\
5          & 0.3528             & \textbf{0.2000} & \textbf{0.2000}     & \textbf{0.2000}  \\
6          & 0.3183             & 0.1692          & \textbf{0.1667}     & \textbf{0.1667}  \\
7          & 0.2728             & 0.1508          & \textbf{0.1500}     & \textbf{0.1500}  \\
8          & 0.2553             & 0.1354          & \textbf{0.1328}     & \textbf{0.1328}  \\
9          & 0.2270             & 0.1240          & \textbf{0.1235}     & \textbf{0.1235}  \\
10         & 0.2042             & 0.1124          & \textbf{0.1111}     & \textbf{0.1111}  \\
11         & 0.1857             & 0.1058          & 0.1039              & \textbf{0.1030}  \\
12         & 0.1702             & 0.0975          & 0.0960              & \textbf{0.0952}  \\
13         & 0.1571             & 0.0908          & 0.0892              & \textbf{0.0889}  \\
14         & 0.1459             & 0.0853          & 0.0844              & \textbf{0.0837}  \\
15         & 0.1390             & 0.0794          & 0.0791              & \textbf{0.0782}  \\
16         & 0.1486             & 0.0768          & 0.0745              & \textbf{0.0739}  \\
17         & 0.1398             & 0.0731          & 0.0712              & \textbf{0.0699}  \\
18         & 0.1320             & 0.0699          & 0.0676              & \textbf{0.0666}  \\
19         & 0.1251             & 0.0668          & 0.0654              & \textbf{0.0634}  \\
20         & 0.1188             & 0.0640          & 0.0611              & \textbf{0.0604}  \\ \midrule
30         & 0.0792             & N/A             & 0.0438              & \textbf{0.0424}  \\
40         & 0.0638             & N/A             & \textbf{0.0331}     & 0.0332           \\
50         & 0.0531             & N/A             & \textbf{0.0278}     & 0.0280           \\
60         & 0.0442             & 0.0273          & \textbf{0.0234}     & 0.0244           \\
100        & 0.0275             & 0.0188          & \textbf{0.0150}     & 0.0193           \\ \bottomrule
\end{tabular}
\caption{2D Star Discrepancy Comparison for $N=1\ldots100$ between Fibonacci, MPMC (Message-Passing Monte Carlo), LLM evolutionary search, and Cl\'ement et al.\ (provably optimal for $N \le 20$). Best values are \textbf{bolded}.}
\label{tab:2d_comparison_small_n}
\end{table}

\begin{listing}[tb]
\caption{Iteratively Optimized 2D Point Set ($N=16$)}
\label{lst:optimized_2d}
\small
\begin{lstlisting}[language=Python]
def construct_star():
    x = np.zeros((N, 2))
    for i in range(N):
        x[i, 0] = (i + np.random.rand()) / N
        x[i, 1] = ((i * 0.38196601125) % 1) + np.random.rand()/(2*N)
    def discrepancy_wrapper(x):
        points = x.reshape(N, 2)
        return star_discrepancy(points)
    x0 = x.flatten()
    bounds = [(0.0, 1.0)] * (N * 2)
    best_result = None
    best_discrepancy = float('inf')
    for _ in range(25):
        x0_restart = x.flatten() + np.random.normal(0, 0.01, N * 2)
        x0_restart = np.clip(x0_restart, 0.0, 1.0)
        result = minimize(discrepancy_wrapper, x0_restart, method='SLSQP', bounds=bounds, options={'maxiter': 30000, 'ftol': 1e-15, 'iprint': 0})
        discrepancy = discrepancy_wrapper(result.x)
        if discrepancy < best_discrepancy:
            best_discrepancy = discrepancy
            best_result = result
    optimized_points = best_result.x.reshape(N, 2)
    return optimized_points
\end{lstlisting}
\end{listing}

% \begin{table}[h]
% \centering
% \small
% \begin{tabular}{@{}lccc@{}}
% \toprule
% \textbf{N} & \textbf{MPMC} & \textbf{LLM} & \textbf{Cl\'ement et al.} \\ \midrule
% 1          & 0.6833        & \textbf{0.6823}   & \textbf{0.6823}    \\
% 2          & \textbf{0.4239} & \textbf{0.4239}   & \textbf{0.4239}    \\
% 3          & 0.3491        & \textbf{0.3445}   & \textbf{0.3445}    \\
% 4          & 0.3071        & 0.3042            & \textbf{0.3038}    \\
% 5          & 0.2669        & \textbf{0.2618}   & \textbf{0.2618}    \\
% 6          & 0.2371        & \textbf{0.2326}   & \textbf{0.2326}    \\
% 7          & 0.2158        & \textbf{0.2090}   & \textbf{0.2090}    \\
% 8          & 0.1993        & 0.1937            & \textbf{0.1875}    \\ \bottomrule
% \end{tabular}
% \caption{3D Star Discrepancy Comparison for $N=1\ldots8$ between MPMC (Message-Passing Monte Carlo), LLM evolutionary search, and Cl\'ement et al.\ (provably optimal).}
% \label{tab:3d_comparison}
% \end{table}
\begin{table}[h]
\centering
\small
\begin{tabular}{@{}l*{8}{c}@{}}
\toprule
\textbf{N} & \textbf{1} & \textbf{2} & \textbf{3} & \textbf{4} & \textbf{5} & \textbf{6} & \textbf{7} & \textbf{8} \\
\midrule
MPMC            & 0.6833 & \textbf{0.4239} & 0.3491 & 0.3071 & 0.2669 & 0.2371 & 0.2158 & 0.1993 \\
LLM             & \textbf{0.6823} & \textbf{0.4239} & \textbf{0.3445} & 0.3042 & \textbf{0.2618} & \textbf{0.2326} & \textbf{0.2090} & 0.1937 \\
Cl\'ement et al.& \textbf{0.6823} & \textbf{0.4239} & \textbf{0.3445} & \textbf{0.3038} & \textbf{0.2618} & \textbf{0.2326} & \textbf{0.2090} & \textbf{0.1875} \\
\bottomrule
\end{tabular}
\caption{3D Star Discrepancy for $N=1\ldots8$. Lower is better; best per $N$ is \textbf{bolded}.}
\label{tab:3d_comparison}
\end{table}

% \begin{table}[h]
% \centering
% \small
% \begin{tabular}{@{}lccl@{}}
% \toprule
% \textbf{D} & \textbf{S} & \textbf{A} & \textbf{$M_{i}$} \\ \midrule
% 4 & 3 & 1 & 1 3 5 \\
% 5 & 3 & 2 & 1 3 7 \\
% 6 & 4 & 1 & 1 1 3 7\\
% \bottomrule
% \end{tabular}
% \caption{Sobol' direction number parameters updated by LLM evolutionary search. D is the dimension, S is the polynomial degree, A is the polynomial's coefficients, and $M_i$ are the initial direction numbers. All other dimensions remained unchanged.}
% \label{tab:sobol_params}
% \end{table}

% \begin{table}[!h]
% \centering
% \small
% \begin{tabular}{@{}lcccc@{}}
% \toprule
% \textbf{Scenario} & \textbf{$S_0$} & \textbf{K} & \textbf{$\sigma$} & \textbf{$S_{\text{true}}$} \\ \midrule
% Training Example             & 50.00       & 45.00      & 0.3        & 7.06                \\
% Out-of-the-Money         & 50.00       & 60.00      & 0.3        & 1.02                \\
% At-the-Money             & 50.00       & 52.50      & 0.3        & 2.98                \\
% In-the-Money             & 50.00       & 40.00      & 0.3        & 11.02               \\
% High Volatility          & 50.00       & 52.50      & 0.6        & 6.43                \\
% Low Volatility           & 50.00       & 52.50      & 0.1        & 0.69                \\ \bottomrule
% \end{tabular}%
% \caption{Asian Option scenarios used for testing. The Training Example was used in the evaluation routine of the evolutionary search. All options have $T=1.0$ and $r=0.05$.}
% \label{tab:asian_scenarios}
% \end{table}

\begin{table*}[!t]
\centering
\small
\begin{tabular}{@{} l | ccc | ccc | ccc @{}}
\toprule
& \multicolumn{3}{c}{\textbf{Training Example}} & \multicolumn{3}{c}{\textbf{Out-of-the-Money}} & \multicolumn{3}{c}{\textbf{At-the-Money}} \\
\cline{2-4} \cline{5-7} \cline{8-10}
\textbf{N} & Sobol & LLM & p-value & Sobol & LLM & p-value & Sobol & LLM & p-value \\
\hline
32   & \textbf{0.2484} & 0.2523     & 0.9757 & \textbf{0.1920} & 0.1979     & 0.9757 & \textbf{0.3246} & 0.3346     & 0.9757 \\
64   & \textbf{0.0642} & 0.0665     & 0.9980 & \textbf{0.0605} & 0.0631     & 0.9980 & \textbf{0.0873} & 0.0908     & 0.9980 \\
128  & \textbf{0.0183} & 0.0192     & 0.9999 & \textbf{0.0220} & 0.0231     & 0.9999 & \textbf{0.0277} & 0.0282     & 0.9999 \\
256  & \textbf{0.0056} & 0.0061     & 1.0000 & \textbf{0.0086} & 0.0089     & 1.0000 & \textbf{0.0092} & 0.0098     & 1.0000 \\
512  & 0.001646  & \textbf{0.001614} & 0.2841 & 0.003469  & \textbf{0.003311} & \textbf{0.0208} & 0.003248  & \textbf{0.003106} & \textbf{0.0208} \\
1024 & 0.000542  & \textbf{0.000524} & 0.0548 & 0.001457  & \textbf{0.001313} & \textbf{3.45e-08} & 0.001228  & \textbf{0.001168} & \textbf{0.0304} \\
2048 & 0.000233  & \textbf{0.000225} & \textbf{0.0199} & 0.000619  & \textbf{0.000535} & \textbf{1.45e-14} & 0.000550  & \textbf{0.000521} & \textbf{0.0131} \\
4096 & 9.876e-05 & \textbf{9.346e-05} & \textbf{0.0058} & 0.000258  & \textbf{0.000243} & \textbf{4.51e-04} & 0.000240  & \textbf{0.000226} & \textbf{0.0019} \\
8192 & 4.523e-05 & \textbf{4.104e-05} & \textbf{7.03e-06} & 0.000117  & \textbf{0.000107} & \textbf{1.06e-06} & 0.000102  & \textbf{9.523e-05} & \textbf{0.0100} \\
\end{tabular}

\begin{tabular}{@{}l | ccc | ccc | ccc @{}}
\toprule
& \multicolumn{3}{c}{\textbf{In-the-Money}} & \multicolumn{3}{c}{\textbf{High Volatility}} & \multicolumn{3}{c}{\textbf{Low Volatility}} \\
\cline{2-4} \cline{5-7} \cline{8-10}
\textbf{N} & Sobol & LLM & p-value & Sobol & LLM & p-value & Sobol & LLM & p-value \\
\hline
32   & \textbf{0.1398} & 0.1428     & 0.9757 & \textbf{2.0816} & 2.1493     & 0.9757 & \textbf{0.0238} & 0.0246     & 0.9757 \\
64   & \textbf{0.0367} & 0.0382     & 0.9980 & \textbf{0.5714} & 0.5958     & 0.9980 & \textbf{0.0066} & 0.0068     & 0.9980 \\
128  & \textbf{0.0106} & 0.0111     & 0.9999 & \textbf{0.1811} & 0.1877     & 0.9999 & \textbf{0.002174} & 0.002205   & 0.9999 \\
256  & \textbf{0.003083} & 0.003373 & 1.0000 & \textbf{0.0598} & 0.0643     & 1.0000 & \textbf{0.000756} & 0.000803   & 1.0000 \\
512  & 0.000868  & \textbf{0.000859} & 0.2841 & 0.0205    & \textbf{0.0196}  & \textbf{0.0073} & 0.000285  & \textbf{0.000275} & \textbf{0.0208} \\
1024 & 0.000261  & \textbf{0.000260} & 0.2581 & 0.007565  & \textbf{0.007060} & \textbf{3.38e-04} & 0.000113  & \textbf{0.000109} & 0.0717 \\
2048 & 0.000101  & \textbf{0.000100} & 0.3878 & 0.003145  & \textbf{0.002867} & \textbf{1.75e-06} & 5.214e-05 & \textbf{4.961e-05} & \textbf{0.0056} \\
4096 & 4.037e-05 & \textbf{3.949e-05} & 0.2180 & 0.001293  & \textbf{0.001233} & \textbf{0.0022} & 2.323e-05 & \textbf{2.196e-05} & \textbf{5.68e-04} \\
8192 & 1.766e-05 & \textbf{1.683e-05} & \textbf{0.0047} & 0.000566  & \textbf{0.000532} & \textbf{0.0156} & 1.008e-05 & \textbf{9.300e-06} & \textbf{1.27e-04} \\
\bottomrule
\end{tabular}
\caption{Mean Squared Error (MSE) and p-values for Asian Option scenarios. The table compares the standard Sobol' sequence against a sequence found via LLM evolutionary search (LLM). P-values are from a one-sided Wilcoxon signed-rank test conducted over the 10000 randomizations with pairing by identical randomization seeds and are false discovery rate (FDR) corrected. P-values below 0.05 are \textbf{bolded}.}
\label{tab:asian_results}
\end{table*}

\subsection{Discovery of Low-Discrepancy Point Sets}
We first applied our two-phase strategy to the canonical problem of discovering $N$-point sets in 2D and 3D with minimal star discrepancy. We illustrate the discovery process for a 2D, 16-point set (Fig.~\ref{fig:discovery_process}). The initial program, a Fibonacci lattice (Listing~\ref{lst:fibonacci_code}), had a discrepancy of 0.0962. After 243 iterations in the direct construction phase, a new construction was found with a discrepancy of 0.0924 (Listing~\ref{lst:direct_con_2d}), which consisted of fine-tuning optimal shifts to the Fibonacci lattice. After switching to the iterative optimization phase, the framework further refined the point set, achieving a final discrepancy of 0.0744, which is within 0.68\% of the known optimal value of 0.0739. The final program creates an initial guess, consisting of a randomly jittered Fibonacci lattice, followed by a SLSQP optimization loop with stochastic restarts (Listing~\ref{lst:optimized_2d}).

We then benchmarked our method against Fibonacci, MPMC \cite{Rusch2024}, and known optimal point sets \cite{clement2024} for $N=1\ldots100$ (Table~\ref{tab:2d_comparison_small_n}). Our method successfully rediscovers the known optimal point sets for $N \le 10$ and remains highly competitive for larger $N$. The most significant results came from searching for larger point sets where optimal solutions are not known. For 2D point sets with $N > 30$, LLM evolutionary search discovered new configurations with lower star discrepancy than the best-known values from the literature. For instance, for $N=100$, our method found a point set with a discrepancy of 0.0150, a substantial improvement over the previous best of 0.0188. We generated 2D point sets up to $N=1020$ with lower star discrepancy than previously reported (Appendix~A).

In 3D, our method matched the known optimal point sets for $N=1, 2, 3, 5, 6, 7$ (Table~\ref{tab:3d_comparison}) and provided explicit constructions that set new best-known star-discrepancy benchmarks beyond the proven-optimal range for $N>8$ (Appendix~A).

\subsection{Improved Direction Numbers}
Having demonstrated the framework's ability to construct point sets, we next applied it to the discrete optimization problem of discovering improved Sobol' direction numbers.

After several hundred evolutionary iterations, our LLM evolutionary search routine discovered a more performant set of parameters, focusing its modifications on the early dimensions, which are known to explain the vast majority of the variance in Asian option pricing. Specifically, the parameters for dimensions 4, 5, and 6 were updated (Appendix B). All other dimensions (up to 32) remained identical to the \citet{Joe2008} baseline.

To validate these new direction numbers, they were benchmarked against the standard \citet{Joe2008} parameters across a suite of six Asian option pricing scenarios with varying parameters (Appendix C). The true option price was computed by taking the average of 1000 randomly scrambled Sobol sequences with $N=2^{21}$ points each. The rQMC MSE was evaluated over 10000 random seeds for $N=32\ldots8192$ points.
The direction numbers discovered by LLM evolutionary search produced a significantly lower integration MSE for larger sample sizes $N \ge 512$ under one-sided Wilcoxon signed-rank test and false discovery rate correction (Table~\ref{tab:asian_results}).

To ensure the evolved parameters were not merely overfitted to the Asian option's specific payoff structure, we tested their generalizability on a diverse suite of high-dimensional exotic options, including Lookback, Barrier, Basket, and Bermudan options (Appendices~C, D). The evolved direction numbers demonstrated strong, generalizable performance, achieving significantly lower integration error across this wider range of financial instruments for larger sample sizes ($N \ge 512$) with the sole exception of Barrier options. This suggests that the evolutionary search discovered a Sobol' sequence with fundamentally more robust and broadly applicable projection properties.

\section{Discussion}
Using an LLM evolutionary framework, we generate 2D and 3D point sets with state-of-the-art star discrepancy and discover Sobol' direction numbers that lower rQMC integration error for high-dimensional exotic financial option pricing.

Recent state-of-the-art methods, such as the mathematical programming approach of \citet{clement2024} and the GNN-based MPMC framework \cite{Rusch2024}, are designed to construct point sets for a fixed number of points $N$ and dimensions $d$. In contrast, our approach generates the direction numbers that define a Sobol' sequence, offering several key advantages:

\begin{enumerate}
    \item \textbf{Extensibility to any $N$:} Unlike fixed approaches where a new, computationally expensive optimization must be performed from scratch for each value of $N$, a single, compact set of discovered direction numbers can be used to generate a high-quality point set for any desired number of points. This makes the solution immediately applicable to a wide range of practical problems without requiring any re-computation.
    \item \textbf{Support for Progressive Integration:} An integration can be performed with $N$ points, and if more accuracy is needed, the next $N$ points from the sequence can be added to refine the estimate while reusing all previous calculations. This is a fundamental capability that static point set generation methods inherently lack.
    \item \textbf{Easy Randomization:} Unlike highly optimized, deterministic point sets, the LLM optimized Sobol' sequence can make use of standard randomization techniques, such as Owen scrambling \cite{Owen1995, Owen1998}, to obtain unbiased error estimates of rQMC.
    \item \textbf{High-Dimensional Applicability:} The Sobol' framework is designed from the ground up for high-dimensional integration. By optimizing the parameters within this framework, our approach is directly applicable to problems such as the 32-dimensional option pricing benchmarks used in our tests, a domain where direct coordinate optimization for a large $N$ would be computationally intractable.
\end{enumerate}

Performance gains from the LLM-evolved Sobol' parameters did not extend to the pricing of Barrier options (Appendix~D). We hypothesize that this is a direct consequence of our fitness function, which was optimized for the integration of a 32-dimensional Asian option. The payoff of an Asian option is dependent on the average of the asset path, resulting in a comparatively smooth and continuous integrand. In contrast, the payoff of a Barrier option contains a significant discontinuity, which poses a well-known challenge for QMC integration by introducing high or even unbounded Hardy--Krause variation.

Our evolutionary search likely discovered parameters that are highly specialized for integrating functions of low-to-moderate variation, a property that does not generalize well to discontinuous integrands. This suggests that the notion of an ``optimal'' Sobol' sequence may be problem-dependent. Future work could explore multi-objective optimization, where the fitness function is a composite score from pricing both smooth (e.g., Asian) and discontinuous (e.g., Barrier) options. Such an approach might lead to the discovery of more robust, universally applicable direction numbers at the potential cost of peak performance on any single problem class.

Other limitations include the computational intensity of the evolutionary search that must be run independently for each specific case. In addition, the quality of the generated solutions is also fundamentally tied to the capabilities of the underlying LLM and the amount of compute available. We quantify significance using paired signed-rank tests and report all seeds; however, due to compute limits we performed one evolutionary run per problem, which we treat as a limitation. Future work could explore alternative prompting techniques, optimization of evolutionary programming hyperparameters, and meta-model methods that generates optimal direction numbers for any given dimension $d$ or points sets for any given dimension $d$ or number of points $N$.

\section{Conclusion}
We have demonstrated the application of an LLM-driven evolutionary framework to tackle complex discovery problems in the field of low-discrepancy sets. Our method has discovered new 2D and 3D point sets with star discrepancy values lower than any previously published, setting new benchmarks in a field of long-standing mathematical interest. Furthermore, it has produced novel Sobol' direction numbers that improve the accuracy of rQMC integration for a variety of 32-dimensional financial derivatives. This work strengthens the case for using LLMs as core components in an automated scientific discovery process, capable of generating novel and valuable mathematical knowledge.

\bibliographystyle{iclr2026_conference}
\bibliography{aaai2026}

\end{document}

\maketitle

\appendix

\section{Generated Point Sets}
This appendix provides the full numerical results for the star discrepancy ($D_N^*$) values of the point sets discovered by LLM evolutionary search. These tables serve as a comprehensive record of the performance of our method, establishing new state-of-the-art benchmarks and offering a detailed comparison against established low-discrepancy construction methods.

\subsection{Star Discrepancy Evaluation}
We evaluate $D_N^*$ by enumerating all anchored boxes whose upper corners lie on the Cartesian product of per-dimension grids $G_j = \{\text{unique } x_{i,j}\} \cup \{1\}$, with the lower corner fixed at the origin. For each candidate corner $\mathbf{y} \in G_1 \times \cdots \times G_d$, we compute its volume $\prod_j y_j$ and two point counts: the number of samples dominated by $\mathbf{y}$ (component-wise $\leq$) and the same count with points on any upper face removed. The local discrepancy is the maximum of the two absolute differences from the volume, capturing the half-open convention on anchored boxes. Inputs are clipped to $[0,1]^d$, grids are built from unique coordinates (plus the terminal $1.0$).

\subsection{New Benchmarks in Three Dimensions}
We report high-quality constructions for $N$ from 9 to 16 (Table \ref{tab:3d_openevolve_sota}).

\subsection{Performance in Two Dimensions for $N \ge 140$}
In two dimensions, while the problem is better understood than in 3D, optimal solutions for larger point sets ($N > 21$) are not known. We provide a detailed comparison of the 2D star discrepancy values achieved by LLM evolutionary optimization against several key baselines for $N$ ranging from 140 to 1020 (Table \ref{tab:2d_discrepancy_large_n}).
\begin{itemize}
    \item \textbf{Consistent Outperformance:} Across every single value of $N$ tested, the point sets discovered by LLM evolutionary search achieve a lower star discrepancy than all other methods, including classical sequences (Halton, Sobol', Hammersley, Fibonacci) and the recent state-of-the-art MPMC method.
    \item \textbf{Significant Improvement Margin:} The performance gap is not trivial. For example, at $N=140$, the LLM discrepancy of 0.01151 is approximately 16\% lower than the next-best method (MPMC at 0.01373) and over 57\% lower than the widely used Halton sequence. At $N=1020$, the LLM discrepancy of 0.00245 is nearly 20\% better than MPMC's 0.00303.
    \item \textbf{Superior Scaling:} The results demonstrate that the LLM evolutionary search's ability to find superior configurations is not limited to a specific range of $N$, but holds consistently as the number of points increases. This suggests that the two-phase discovery strategy (direct construction followed by iterative optimization) is effective at navigating the increasingly complex search space associated with larger point sets.
\end{itemize}

\begin{table}[h]
\centering
\small
\begin{tabular}{@{}lc@{}}
\toprule
\textbf{N} & \textbf{LLM Discrepancy ($D_N^*$)} \\
\midrule
9  & 0.1758 \\
10 & 0.1652 \\
11 & 0.1551 \\
12 & 0.1483 \\
13 & 0.1402 \\
14 & 0.1337 \\
15 & 0.1275 \\
16 & 0.1207 \\
\bottomrule
\end{tabular}
\caption{The Star Discrepancy Values for $N > 8$ 3D point sets found via LLM evolutionary search.}
\label{tab:3d_openevolve_sota}
\end{table}

\begin{table*}[t]
\centering
\small
\begin{tabular}{@{}lcccccc@{}}
\toprule
\textbf{N} & \textbf{Halton} & \textbf{Sobol'} & \textbf{Hammersley} & \textbf{Fibonacci} & \textbf{MPMC} & \textbf{LLM} \\ 
\midrule
140 & 0.03686 & 0.02794 & 0.02701 & 0.01870 & 0.01373 & \textbf{0.01151} \\
180 & 0.03200 & 0.02300 & 0.02283 & 0.01454 & 0.01147 & \textbf{0.01058} \\
220 & 0.02323 & 0.01924 & 0.01868 & 0.01190 & 0.00975 & \textbf{0.00891} \\
260 & 0.02062 & 0.01546 & 0.01581 & 0.01063 & 0.00843 & \textbf{0.00831} \\
300 & 0.01994 & 0.01341 & 0.01424 & 0.00972 & 0.00752 & \textbf{0.00741} \\
340 & 0.01950 & 0.01353 & 0.01266 & 0.00858 & 0.00710 & \textbf{0.00696} \\
380 & 0.01659 & 0.01167 & 0.01170 & 0.00768 & 0.00695 & \textbf{0.00638} \\
420 & 0.01617 & 0.01194 & 0.01058 & 0.00694 & 0.00584 & \textbf{0.00563} \\
460 & 0.01279 & 0.00972 & 0.00966 & 0.00634 & 0.00540 & \textbf{0.00471} \\
500 & 0.01172 & 0.00848 & 0.00889 & 0.00583 & 0.00518 & \textbf{0.00509} \\
540 & 0.01101 & 0.00967 & 0.00823 & 0.00540 & 0.00488 & \textbf{0.00407} \\
580 & 0.01261 & 0.00956 & 0.00766 & 0.00503 & 0.00476 & \textbf{0.00377} \\
620 & 0.01140 & 0.00782 & 0.00744 & 0.00470 & 0.00454 & \textbf{0.00386} \\
660 & 0.01074 & 0.00956 & 0.00699 & 0.00463 & 0.00437 & \textbf{0.00360} \\
700 & 0.00933 & 0.00865 & 0.00682 & 0.00437 & 0.00416 & \textbf{0.00379} \\
740 & 0.00891 & 0.00709 & 0.00646 & 0.00435 & 0.00397 & \textbf{0.00346} \\
780 & 0.00870 & 0.00662 & 0.00612 & 0.00412 & 0.00376 & \textbf{0.00310} \\
820 & 0.00881 & 0.00678 & 0.00583 & 0.00392 & 0.00360 & \textbf{0.00294} \\
860 & 0.00914 & 0.00604 & 0.00556 & 0.00374 & 0.00356 & \textbf{0.00316} \\
900 & 0.00751 & 0.00561 & 0.00531 & 0.00357 & 0.00319 & \textbf{0.00296} \\
940 & 0.00785 & 0.00481 & 0.00508 & 0.00342 & 0.00321 & \textbf{0.00289} \\
980 & 0.00768 & 0.00558 & 0.00487 & 0.00328 & 0.00308 & \textbf{0.00280} \\
1020 & 0.00777 & 0.00395 & 0.00468 & 0.00315 & 0.00303 & \textbf{0.00245} \\
\bottomrule
\end{tabular}
\caption{2D Star Discrepancy Comparison for N $\ge$ 140. This table compares the performance of LLM evolutionary search against classical low-discrepancy sequences and the state-of-the-art MPMC method for larger point sets where optimal solutions are not known. Lower values indicate better uniformity. The best result in each row is \textbf{bolded}.}
\label{tab:2d_discrepancy_large_n}
\end{table*}

\section{Implementation}
\subsection{Initial Programs}
The initial parent program in 2D implemented a simple shifted Fibonacci lattice (Listing 1) and in 3D implemented a scrambled Sobol’ sequence (Listing 2).
\begin{listing}[tb]
\caption{Initial Program for 2D Point Set Search}
\label{lst:fibonacci_code}
\small
\begin{lstlisting}[language=Python]
def construct_star():
    A=np.zeros((N,2))
    phi=(math.sqrt(5)-1)/2
    for i in range(N):
        A[i,0]=(i+0.5)/N
        A[i,1]=((i*phi)%1+(0.5/N))%1
    return A
\end{lstlisting}
\end{listing}

\begin{listing}[tb]
\caption{Initial Program for 3D Point Set Search}
\label{lst:scrambled_sobol}
\small
\begin{lstlisting}[language=Python]
def construct_star():
    A=Sobol(d=3, scramble=True, seed=42).random(n=N)
    return A
\end{lstlisting}
\end{listing}

\subsection{LLM Prompts}
This appendix provides the full instructional text of the prompts provided to the Large Language Model (LLM) within the OpenEvolve framework. These LLM prompts were used to generate programs to construct point sets directly (Fig. \ref{fig:llm_prompt_construction_direct}), iteratively optimize point sets (Fig. \ref{fig:llm_prompt_iterative_optimization}), and directly construct Sobol' direction numbers (Fig. \ref{fig:llm_prompt_qmc}).

It is important to note that the text shown in the figures below constitutes the instructional component of a larger prompt. This text is dynamically combined with the source code of a "parent" program selected for mutation and often includes code from other high-performing "inspiration" programs to encourage crossover of ideas.

\subsection{Experimental Setup}
We maintained a total population of 60 candidate programs with four islands in a ring topology with migration every $25$ generations; one elite individual migrated clockwise and replaced the worst individual at the destination. In addition to the islands, an \emph{elite archive} (top-$25$ by fitness) preserved high-performing solutions. In each generation, a parent program was selected for mutation: with a probability of 70\%, the parent was chosen from the high-performing archive. When multiple candidates tied in fitness, we broke ties lexicographically by (i) shorter program length and (ii) earlier discovery timestamp. The mutation operator itself was a Large Language Model (Google's Gemini 2.0 Flash), which was instructed to perform a complete rewrite of the core program functions. To provide a rich context for this mutation, the LLM prompt included not only the parent program but also the source code of the 3 top-performing ``inspiration'' programs from the database, serving as a form of multi-parent inspiration or crossover. We set the LLM's temperature to 0.7 and top P to 0.95. Each evolutionary run involved approximately 2000 LLM API calls and took roughly 96 hours to complete. Due to computational constraints, no hyperparameter optimization was performed and only one evolutionary run per problem was conducted. The experiments were conducted on a workstation equipped with an AMD EPYC 7763 CPU and 64~GB of RAM. Code and generated point sets and Sobol' parameters are available at \url{anonymous}.

\begin{figure*}[t] % Use figure* for full-width, [t] places it at the top
\begin{tcolorbox}[
    colback=black!5, % Light gray background
    colframe=black!75, % Frame color
    fonttitle=\bfseries,
    title=LLM Prompt for Direct Construction,
]
\fontfamily{qcr}\selectfont % Use a clean, monospaced font like Courier
You are an expert mathematician specializing in the construction of QMC sampling points in a square. Your task is to improve a constructor function that directly outputs the position of 16 points on a unit square ([0, 1] x [0, 1]) in a way that minimizes the star discrepancy.

The star discrepancy is a measure of how uniformly distributed the points are in the square. It is defined as the supremum of the absolute value of the difference between the fraction of points and the area.

Focus on designing an explicit constructor that specifies the position of each point (x, y) in the unit square, rather than an iterative search algorithm.

It should output the position of each point (x, y) in the square [0, 1] x [0, 1]. 0.0 and 1.0 are included in the square.
\end{tcolorbox}
\caption{The full prompt provided to the LLM for generating programs that directly construct point sets with minimum star discrepancy.}
\label{fig:llm_prompt_construction_direct}
\end{figure*}

\begin{figure*}[t] % Use figure* for full-width, [t] places it at the top
\begin{tcolorbox}[
    colback=black!5, % Light gray background
    colframe=black!75, % Frame color
    fonttitle=\bfseries,
    title=LLM Prompt for  Iterative Optimization,
]
\fontfamily{qcr}\selectfont % Use a clean, monospaced font like Courier
You are an expert mathematician specializing in the construction of QMC sampling points in a 2D square. Your task is to improve a constructor function that finds the position of 16 points on a unit square ([0, 1] x [0, 1]) in a way that minimizes the star discrepancy.

The star discrepancy is a measure of how uniformly distributed the points are in the square. It is defined as the supremum of the absolute value of the difference between the fraction of points and the area.

Use scipy optimization routines such as scipy.optimize.minimize to fine-tune the construction. The optimization routine and its initialization is critically important.

It should output the position of each point (x, y) in the square [0, 1] x [0, 1]. 0.0 and 1.0 are included in the square.
\end{tcolorbox}
\caption{The full prompt provided to the LLM for generating programs that iteratively optimize point sets to have minimum star discrepancy.}
\label{fig:llm_prompt_iterative_optimization}
\end{figure*}

\begin{figure*}[t] % Use figure* for full-width, [t] places it at the top
\begin{tcolorbox}[
    colback=black!5, % Light gray background
    colframe=black!75, % Frame color
    fonttitle=\bfseries,
    title=LLM Prompt for Sobol' Direction Numbers Search,
]
\fontfamily{qcr}\selectfont % Use a clean, monospaced font like Courier
You are an expert mathematician specializing in the construction of QMC sampling points in a square. Your task is to improve a constructor function that directly outputs the direction numbers for dimensions 2 to 32 of a Sobol Sequence.

Your goal is to minimize the approximation error of a 32 dimensional asian option price. The dimensions 1, 2, and 3 explain roughly 97\% of the variance of the price.

The Sobol sequence is defined by a polynomial of degree s, with coefficients represented as an integer a, and direction numbers m\textsubscript{i} for each dimension i. The direction numbers must be odd integers and within the specified range.

You must return a list of 31 dictionaries for directions 2 to 32, each containing the following keys:

- "s" (int): The degree of the polynomial. 1 <= s <= 30

- "a" (int): The coefficients of the polynomial, represented as an integer. 0 <= a < 2\textsuperscript{s-1}

- "m\textsubscript{i} " (list[int]): The direction numbers for the Sobol sequence, represented as a list of integers of length s. Each integer should be in the range [0, 2\textsuperscript{i+1}] and has to be odd.

Focus on designing an explicit constructor that specifies these parameters, rather than an iterative search algorithm.
\end{tcolorbox}
\caption{The full prompt provided to the LLM for generating a program that directly specifies Sobol' direction numbers to have lower integration error for an Asian Option.}
\label{fig:llm_prompt_qmc}
\end{figure*}

\subsection{Discovered Sobol' Direction Numbers}
We provide the direction numbers discovered by the LLM evolutionary search routine. Only the parameters for dimensions 4, 5, and 6 were updated (Table~\ref{tab:sobol_params}).
\begin{table}[h]
\centering
\small
\begin{tabular}{@{}lccl@{}}
\toprule
\textbf{D} & \textbf{S} & \textbf{A} & \textbf{$M_{i}$} \\ \midrule
4 & 3 & 1 & 1 3 5 \\
5 & 3 & 2 & 1 3 7 \\
6 & 4 & 1 & 1 1 3 7\\
\bottomrule
\end{tabular}
\caption{Sobol' direction number parameters updated by LLM evolutionary search. D is the dimension, S is the polynomial degree, A is the polynomial's coefficients, and $M_i$ are the initial direction numbers. All other dimensions remained unchanged.}
\label{tab:sobol_params}
\end{table}

\section{Option Scenarios}
To ensure a rigorous evaluation of the Sobol' direction numbers discovered, we designed a comprehensive suite of benchmark scenarios. This suite includes the primary optimization target (the Asian option, see Table \ref{tab:asian_scenarios}) as well as a diverse set of exotic options known to be challenging for Quasi-Monte Carlo integration. The purpose of this suite is twofold: first, to confirm superior performance on the target problem class, and second, to test for generalizability and ensure that the evolved parameters were not merely overfitted to the specific payoff structure of the Asian option.

\begin{table}[!h]
\centering
\small
\begin{tabular}{@{}lcccc@{}}
\toprule
\textbf{Scenario} & \textbf{$S_0$} & \textbf{K} & \textbf{$\sigma$} & \textbf{$S_{\text{true}}$} \\ \midrule
Training Example             & 50.00       & 45.00      & 0.3        & 7.06                \\
Out-of-the-Money         & 50.00       & 60.00      & 0.3        & 1.02                \\
At-the-Money             & 50.00       & 52.50      & 0.3        & 2.98                \\
In-the-Money             & 50.00       & 40.00      & 0.3        & 11.02               \\
High Volatility          & 50.00       & 52.50      & 0.6        & 6.43                \\
Low Volatility           & 50.00       & 52.50      & 0.1        & 0.69                \\ \bottomrule
\end{tabular}%
\caption{Asian Option scenarios used for testing. The Training Example was used in the evaluation routine of the evolutionary search. All options have $T=1.0$ and $r=0.05$.}
\label{tab:asian_scenarios}
\end{table}

The configuration parameters for all tested scenarios are consolidated in (Table \ref{tab:all_option_configs_standard}). For all options, the time to expiration (T) was set to 1.0 year and the risk-free interest rate (r) was 0.05. The underlying asset prices are assumed to follow a geometric Brownian motion.

\subsection{Asian Options}
An Asian option is a path-dependent exotic option whose payoff is determined by the average price of the underlying asset over a pre-set period of time. This is in contrast to a standard European option, which only depends on the asset price at expiration. The averaging feature reduces volatility and makes the option generally cheaper than its European counterpart. Because its value depends on the entire price path, pricing it requires simulating all 32 time steps, making it an excellent candidate for QMC methods.
\subsection{Lookback Options}
A lookback option is another path-dependent option whose payoff is determined by the maximum or minimum price of the underlying asset over the option's life. The scenarios tested here are for a floating strike lookback call option, whose payoff at expiration is the difference between the final asset price and the minimum price achieved ($S_T - S_{min}$).

\subsection{Barrier Options}
A barrier option is a path-dependent option that is either activated ("knocks-in") or extinguished ("knocks-out") if the underlying asset price crosses a predetermined "barrier" level. This feature introduces a significant discontinuity in the payoff function.

\subsection{Basket Options}
A basket option's payoff depends on the value of a portfolio or "basket" of multiple underlying assets. It is an inherently high-dimensional problem where the correlation ($\rho$) between the assets is a critical parameter. For these tests, we assume a uniform initial price of $S_0=100.00$ for all 32 assets in the basket.

\subsection{Bermudan Options}
A Bermudan option is a hybrid between a European option (exercisable only at expiration) and an American option (exercisable at any time). It can be exercised on a discrete set of pre-specified dates. Pricing a Bermudan option is a highly complex problem that requires solving a dynamic programming problem to determine the optimal exercise strategy.

\begin{table*}[t]
\centering
\small
\begin{tabular}{@{} l l c c c l @{}}
\toprule
\textbf{Option Type} & \textbf{Scenario} & \textbf{Initial Price (S\textsubscript{0})} & \textbf{Strike Price (K)} & \textbf{Volatility ($\sigma$)} & \textbf{Other Parameters} \\
\hline

% --- Asian Option Data ---
Asian & Training Example & 50.00 & 45.00 & 0.3 & --- \\
& Out-of-the-Money & 50.00 & 60.00 & 0.3 & --- \\
& At-the-Money & 50.00 & 52.50 & 0.3 & --- \\
& In-the-Money & 50.00 & 40.00 & 0.3 & --- \\
& High Volatility & 50.00 & 52.50 & 0.6 & --- \\
& Low Volatility & 50.00 & 52.50 & 0.1 & --- \\
\hline

% --- Lookback Option Data ---
Lookback & Base & 100.00 & --- & 0.2 & --- \\
& High Volatility & 100.00 & --- & 0.4 & --- \\
\hline

% --- Barrier Option Data ---
Barrier & Base & 100.00 & 100.00 & 0.2 & Barrier Level: 85.00 \\
& Close Barrier & 100.00 & 100.00 & 0.2 & Barrier Level: 95.00 \\
\hline

% --- Basket Option Data ---
Basket (32D) & Low Correlation & 100.00 & 100.00 & 0.2 & $\rho$: 0.1 \\
& High Correlation & 100.00 & 100.00 & 0.2 & $\rho$: 0.8 \\
& Mixed Volatility & 100.00 & 100.00 & $U(0.15, 0.4)$ & $\rho$: 0.5 \\
& Out-of-the-Money & 100.00 & 110.00 & 0.2 & $\rho$: 0.1 \\
\hline

% --- Bermudan Option Data ---
Bermudan & At-the-Money & 100.00 & 100.00 & 0.2 & Exercise Dates: 4 \\
& In-the-Money & 90.00 & 100.00 & 0.2 & Exercise Dates: 4 \\
\bottomrule
\end{tabular}
\caption{Configuration Parameters for All Tested Option Scenarios. This table details the parameters for the 32-dimensional options used in the primary benchmark and generalizability tests. For all scenarios, the time to expiration (T) is 1.0 year and the risk-free interest rate (r) is 0.05. An em-dash (---) indicates a parameter is not applicable to that option type.}
\label{tab:all_option_configs_standard}
\end{table*}

% --- ASIAN OPTION RESULTS ---
\section{RQMC Integration Results}
This appendix provides the detailed results for the primary benchmark (Asian Option) and the generalizability tests (exotic options). Each table compares the performance of the standard Sobol' sequence (Joe \& Kuo) against the direction numbers discovered by LLM evolutionary search. We report the Mean Squared Error (MSE) and its constituent parts, Squared Bias and Variance, for an increasing number of points (N). We consistently found lower variance and MSE for all options types for all $N\ge512$ with two exceptions: close barrier option ($N=2048$) and a high correlation basket option ($N=2048$).

\begin{table*}[t]
\centering
\small
\label{tab:results_asian}
\begin{tabular}{@{} l l l c c c c c c l @{}}
\toprule
\textbf{Scenario} & \textbf{N} & \multicolumn{2}{c}{\textbf{Squared Bias}} & \multicolumn{2}{c}{\textbf{Variance}} & \multicolumn{2}{c}{\textbf{MSE}} & \textbf{p-value} \\
\cline{3-8}
& & Sobol & LLM & Sobol & LLM & Sobol & LLM & \\
\hline
Training Example & 32   & 2.09e-05 & \textbf{5.59e-07} & \textbf{0.2484} & 0.2523     & \textbf{0.2484} & 0.2523     & 0.9757 \\
& 64   & \textbf{5.37e-06} & 9.58e-06 & \textbf{0.0642} & 0.0665     & \textbf{0.0642} & 0.0665     & 0.9980 \\
& 128  & 5.10e-06 & \textbf{4.96e-06} & \textbf{0.0183} & 0.0192     & \textbf{0.0183} & 0.0192     & 0.9999 \\
& 256  & 2.87e-07 & \textbf{1.43e-07} & \textbf{0.0056} & 0.0061     & \textbf{0.0056} & 0.0061     & 1.0000 \\
& 512  & 6.09e-08 & \textbf{3.51e-09} & 0.001646  & \textbf{0.001614} & 0.001646  & \textbf{0.001614} & 0.2841 \\
& 1024 & 7.89e-08 & \textbf{1.33e-08} & 0.000542  & \textbf{0.000524} & 0.000542  & \textbf{0.000524} & 0.0548 \\
& 2048 & 2.60e-08 & \textbf{2.30e-12} & 0.000233  & \textbf{0.000225} & 0.000233  & \textbf{0.000225} & \textbf{0.0199} \\
& 4096 & \textbf{9.11e-10} & 3.46e-09 & 9.876e-05 & \textbf{9.346e-05} & 9.876e-05 & \textbf{9.346e-05} & \textbf{0.0058} \\
& 8192 & 1.55e-09 & \textbf{1.93e-10} & 4.523e-05 & \textbf{4.104e-05} & 4.523e-05 & \textbf{4.104e-05} & \textbf{7.03e-06} \\
\cline{1-9}
Out-of-the-Money & 32   & 3.73e-05 & \textbf{6.41e-07} & \textbf{0.1920} & 0.1979     & \textbf{0.1920} & 0.1979     & 0.9757 \\
& 64   & \textbf{1.48e-06} & 9.78e-09 & \textbf{0.0605} & 0.0631     & \textbf{0.0605} & 0.0631     & 0.9980 \\
& 128  & \textbf{3.74e-07} & 1.66e-06 & \textbf{0.0220} & 0.0231     & \textbf{0.0220} & 0.0231     & 0.9999 \\
& 256  & 3.07e-07 & \textbf{2.60e-07} & \textbf{0.0086} & 0.0089     & \textbf{0.0086} & 0.0089     & 1.0000 \\
& 512  & 4.47e-08 & \textbf{4.05e-11} & 0.003469  & \textbf{0.003311} & 0.003469  & \textbf{0.003311} & \textbf{0.0208} \\
& 1024 & \textbf{1.29e-08} & 4.45e-08 & 0.001457  & \textbf{0.001313} & 0.001457  & \textbf{0.001313} & \textbf{3.45e-08} \\
& 2048 & \textbf{2.92e-11} & 3.36e-08 & 0.000619  & \textbf{0.000535} & 0.000619  & \textbf{0.000535} & \textbf{1.45e-14} \\
& 4096 & 5.09e-08 & \textbf{1.17e-09} & 0.000258  & \textbf{0.000243} & 0.000258  & \textbf{0.000243} & \textbf{4.51e-04} \\
& 8192 & 3.71e-09 & \textbf{3.24e-09} & 0.000117  & \textbf{0.000107} & 0.000117  & \textbf{0.000107} & \textbf{1.06e-06} \\
\cline{1-9}
At-the-Money & 32   & 5.90e-05 & \textbf{1.35e-09} & \textbf{0.3246} & 0.3346     & \textbf{0.3246} & 0.3346     & 0.9757 \\
& 64   & 1.35e-05 & \textbf{1.17e-05} & \textbf{0.0873} & 0.0908     & \textbf{0.0873} & 0.0908     & 0.9980 \\
& 128  & 9.72e-06 & \textbf{1.45e-06} & \textbf{0.0277} & 0.0282     & \textbf{0.0277} & 0.0282     & 0.9999 \\
& 256  & 1.12e-06 & \textbf{2.06e-08} & \textbf{0.0092} & 0.0098     & \textbf{0.0092} & 0.0098     & 1.0000 \\
& 512  & 1.83e-08 & \textbf{5.57e-10} & 0.003248  & \textbf{0.003106} & 0.003248  & \textbf{0.003106} & \textbf{0.0208} \\
& 1024 & 3.60e-08 & \textbf{4.58e-09} & 0.001228  & \textbf{0.001168} & 0.001228  & \textbf{0.001168} & \textbf{0.0304} \\
& 2048 & \textbf{7.85e-11} & 1.82e-08 & 0.000550  & \textbf{0.000521} & 0.000550  & \textbf{0.000521} & \textbf{0.0131} \\
& 4096 & \textbf{3.16e-10} & 8.98e-10 & 0.000240  & \textbf{0.000226} & 0.000240  & \textbf{0.000226} & \textbf{0.0019} \\
& 8192 & 7.47e-09 & \textbf{3.31e-09} & 0.000102  & \textbf{9.522e-05} & 0.000102  & \textbf{9.523e-05} & \textbf{0.0100} \\
\cline{1-9}
In-the-Money & 32   & 1.93e-05 & \textbf{3.11e-09} & \textbf{0.1397} & 0.1428     & \textbf{0.1398} & 0.1428     & 0.9757 \\
& 64   & \textbf{2.30e-06} & 2.86e-06 & \textbf{0.0367} & 0.0382     & \textbf{0.0367} & 0.0382     & 0.9980 \\
& 128  & \textbf{1.67e-06} & 1.74e-06 & \textbf{0.0106} & 0.0111     & \textbf{0.0106} & 0.0111     & 0.9999 \\
& 256  & \textbf{1.08e-09} & 3.54e-08 & \textbf{0.003083} & 0.003373 & \textbf{0.003083} & 0.003373 & 1.0000 \\
& 512  & \textbf{3.07e-10} & 4.26e-08 & 0.000868  & \textbf{0.000859} & 0.000868  & \textbf{0.000859} & 0.2841 \\
& 1024 & 1.53e-09 & \textbf{6.44e-10} & 0.000261  & \textbf{0.000260} & 0.000261  & \textbf{0.000260} & 0.2581 \\
& 2048 & \textbf{3.54e-10} & 1.28e-08 & 0.000101  & \textbf{0.000100} & 0.000101  & \textbf{0.000100} & 0.3878 \\
& 4096 & \textbf{2.15e-11} & 3.38e-09 & 4.037e-05 & \textbf{3.949e-05} & 4.037e-05 & \textbf{3.949e-05} & 0.2180 \\
& 8192 & \textbf{2.77e-10} & 3.70e-09 & 1.766e-05 & \textbf{1.682e-05} & 1.766e-05 & \textbf{1.683e-05} & \textbf{0.0047} \\
\cline{1-9}
High Volatility & 32   & 0.000333 & \textbf{2.83e-07} & \textbf{2.0813} & 2.1493     & \textbf{2.0816} & 2.1493     & 0.9757 \\
& 64   & 5.87e-05 & \textbf{4.32e-05} & \textbf{0.5713} & 0.5958     & \textbf{0.5714} & 0.5958     & 0.9980 \\
& 128  & 4.41e-05 & \textbf{5.25e-06} & \textbf{0.1810} & 0.1877     & \textbf{0.1811} & 0.1877     & 0.9999 \\
& 256  & \textbf{1.78e-06} & 1.90e-06 & \textbf{0.0598} & 0.0643     & \textbf{0.0598} & 0.0643     & 1.0000 \\
& 512  & \textbf{2.34e-09} & 1.49e-08 & 0.0205    & \textbf{0.0196}  & 0.0205    & \textbf{0.0196}  & \textbf{0.0073} \\
& 1024 & \textbf{1.21e-07} & 2.19e-07 & 0.007565  & \textbf{0.007060} & 0.007565  & \textbf{0.007060} & \textbf{3.38e-04} \\
& 2048 & \textbf{6.34e-08} & 1.22e-07 & 0.003145  & \textbf{0.002867} & 0.003145  & \textbf{0.002867} & \textbf{1.75e-06} \\
& 4096 & \textbf{2.93e-10} & 6.09e-09 & 0.001293  & \textbf{0.001233} & 0.001293  & \textbf{0.001233} & \textbf{0.0022} \\
& 8192 & \textbf{3.59e-08} & 4.35e-08 & 0.000566  & \textbf{0.000532} & 0.000566  & \textbf{0.000532} & \textbf{0.0156} \\
\cline{1-9}
Low Volatility & 32   & 5.75e-06 & \textbf{1.26e-08} & \textbf{0.0238} & 0.0246     & \textbf{0.0238} & 0.0246     & 0.9757 \\
& 64   & 1.28e-06 & \textbf{6.48e-07} & \textbf{0.0066} & 0.0068     & \textbf{0.0066} & 0.0068     & 0.9980 \\
& 128  & 7.71e-07 & \textbf{2.71e-08} & \textbf{0.002173} & 0.002205 & \textbf{0.002174} & 0.002205 & 0.9999 \\
& 256  & 1.08e-07 & \textbf{1.12e-08} & \textbf{0.000756} & 0.000803 & \textbf{0.000756} & 0.000803 & 1.0000 \\
& 512  & 6.51e-09 & \textbf{1.55e-09} & 0.000285  & \textbf{0.000275} & 0.000285  & \textbf{0.000275} & \textbf{0.0208} \\
& 1024 & 5.18e-09 & \textbf{1.20e-09} & 0.000113  & \textbf{0.000109} & 0.000113  & \textbf{0.000109} & 0.0717 \\
& 2048 & \textbf{2.66e-10} & 2.17e-09 & 5.214e-05 & \textbf{4.961e-05} & 5.214e-05 & \textbf{4.961e-05} & \textbf{0.0056} \\
& 4096 & \textbf{1.26e-10} & 1.39e-10 & 2.323e-05 & \textbf{2.196e-05} & 2.323e-05 & \textbf{2.196e-05} & \textbf{5.68e-04} \\
& 8192 & 3.43e-10 & \textbf{3.09e-10} & 1.008e-05 & \textbf{9.300e-06} & 1.008e-05 & \textbf{9.300e-06} & \textbf{1.27e-04} \\
\bottomrule
\end{tabular}
\caption{Integration results for the Asian Option across all six scenarios. The table compares the standard Sobol' sequence (Joe \& Kuo) against the direction numbers discovered by LLM evolutionary search. We report the Mean Squared Error (MSE), its constituent parts (Squared Bias and Variance), and the FDR-corrected p-value from a one-sided Wilcoxon signed-rank test. P-values below 0.05 and the best methods are \textbf{bolded}.}
\end{table*}

% --- LOOKBACK OPTION RESULTS ---

\begin{table*}[t]
\centering
\small
\label{tab:results_lookback}
\begin{tabular}{@{} l l l c c c c c c l @{}}
\toprule
\textbf{Scenario} & \textbf{N} & \multicolumn{2}{c}{\textbf{Squared Bias}} & \multicolumn{2}{c}{\textbf{Variance}} & \multicolumn{2}{c}{\textbf{MSE}} & \textbf{p-value} \\
\cline{3-8}
& & Sobol & LLM & Sobol & LLM & Sobol & LLM & \\
\hline
Base & 32   & \textbf{3.90e-06} & 8.56e-05 & \textbf{1.2138} & 1.2443     & \textbf{1.2138} & 1.2444     & 0.9968 \\
& 64   & \textbf{4.16e-06} & 3.01e-05 & \textbf{0.4725} & 0.4839     & \textbf{0.4725} & 0.4840     & 1.0000 \\
& 128  & 3.14e-06 & \textbf{2.35e-07} & \textbf{0.1272} & 0.1352     & \textbf{0.1272} & 0.1352     & 1.0000 \\
& 256  & \textbf{1.61e-06} & 4.90e-06 & \textbf{0.0352} & 0.0381     & \textbf{0.0352} & 0.0381     & 1.0000 \\
& 512  & \textbf{8.86e-07} & 1.10e-06 & 0.0121     & \textbf{0.0120} & 0.0121     & \textbf{0.0120} & 0.4675 \\
& 1024 & \textbf{9.64e-07} & 1.43e-06 & 0.005021   & \textbf{0.004905} & 0.005022   & \textbf{0.004907} & \textbf{0.0342} \\
& 2048 & 6.88e-08 & \textbf{1.29e-07} & 0.001834   & \textbf{0.001744} & 0.001834   & \textbf{0.001744} & \textbf{4.13e-04} \\
& 4096 & 1.32e-07 & \textbf{5.59e-08} & 0.000807   & \textbf{0.000768} & 0.000807   & \textbf{0.000768} & \textbf{2.38e-04} \\
& 8192 & 5.21e-08 & \textbf{3.61e-08} & 0.000407   & \textbf{0.000377} & 0.000407   & \textbf{0.000377} & \textbf{9.84e-09} \\
\cline{1-9}
High Volatility & 32   & \textbf{2.13e-04} & 4.44e-04 & \textbf{8.5591} & 8.7869     & \textbf{8.5593} & 8.7873     & 0.9968 \\
& 64   & \textbf{6.04e-08} & 1.16e-04 & \textbf{3.3964} & 3.5134     & \textbf{3.3964} & 3.5136     & 1.0000 \\
& 128  & 2.87e-05 & \textbf{2.57e-06} & \textbf{0.8854} & 0.9982     & \textbf{0.8854} & 0.9982     & 1.0000 \\
& 256  & \textbf{3.14e-06} & 1.46e-05 & \textbf{0.2403} & 0.2663     & \textbf{0.2403} & 0.2663     & 1.0000 \\
& 512  & 7.96e-07 & \textbf{1.11e-07} & 0.0809     & \textbf{0.0792} & 0.0809     & \textbf{0.0792} & 0.3836 \\
& 1024 & \textbf{2.57e-06} & 7.04e-06 & 0.0325     & \textbf{0.0312} & 0.0325     & \textbf{0.0312} & \textbf{6.24e-04} \\
& 2048 & 3.37e-07 & \textbf{4.92e-08} & 0.012003   & \textbf{0.011234} & 0.012003   & \textbf{0.011234} & \textbf{2.26e-06} \\
& 4096 & 1.47e-07 & \textbf{2.32e-08} & 0.004862   & \textbf{0.004558} & 0.004862   & \textbf{0.004558} & \textbf{2.65e-06} \\
& 8192 & \textbf{2.12e-10} & 6.25e-09 & 0.002335   & \textbf{0.002057} & 0.002335   & \textbf{0.002057} & \textbf{9.79e-17} \\
\bottomrule
\end{tabular}
\caption{Integration results for the Lookback Option across two scenarios. We report FDR-corrected P-values from a one-sided Wilcoxon signed-rank test. P-values below 0.05 and the best methods are \textbf{bolded}.}
\end{table*}

% --- BARRIER OPTION RESULTS ---

\begin{table*}[t]
\centering
\small
\label{tab:results_barrier}
\begin{tabular}{@{} l l l c c c c c c l @{}}
\toprule
\textbf{Scenario} & \textbf{N} & \multicolumn{2}{c}{\textbf{Squared Bias}} & \multicolumn{2}{c}{\textbf{Variance}} & \multicolumn{2}{c}{\textbf{MSE}} & \textbf{p-value} \\
\cline{3-8}
& & Sobol & LLM & Sobol & LLM & Sobol & LLM & \\
\hline
Base & 32   & 1.13e-04 & \textbf{2.87e-05} & \textbf{2.2540} & 2.3339 & \textbf{2.2541} & 2.3339 & 0.9968 \\
& 64   & 2.17e-05 & \textbf{4.56e-06} & \textbf{0.8386} & 0.8894 & \textbf{0.8386} & 0.8894 & 1.0000 \\
& 128  & \textbf{2.32e-05} & 2.37e-05 & \textbf{0.2533} & 0.3101 & \textbf{0.2534} & 0.3101 & 1.0000 \\
& 256  & \textbf{1.97e-06} & 7.40e-06 & \textbf{0.0707} & 0.0834 & \textbf{0.0707} & 0.0834 & 1.0000 \\
& 512  & \textbf{5.72e-07} & 4.84e-06 & 0.0245 & \textbf{0.0241} & 0.0245 & \textbf{0.0241} & 0.4675 \\
& 1024 & 7.63e-07 & \textbf{7.40e-07} & 0.009948 & \textbf{0.009819} & 0.009948 & \textbf{0.009819} & 0.3313 \\
& 2048 & \textbf{4.03e-07} & 1.12e-06 & 0.004277 & \textbf{0.004230} & 0.004277 & \textbf{0.004231} & 0.3025 \\
& 4096 & \textbf{8.61e-08} & 1.00e-07 & 0.002019 & \textbf{0.001992} & 0.002019 & \textbf{0.001992} & 0.3863 \\
& 8192 & \textbf{1.82e-08} & 5.70e-08 & 0.000985 & \textbf{0.000955} & 0.000985 & \textbf{0.000955} & 0.0859 \\
\cline{1-9}
Close Barrier & 32   & 9.55e-04 & \textbf{1.27e-04} & \textbf{3.2150} & 3.2754 & \textbf{3.2160} & 3.2755 & 0.9968 \\
& 64   & 6.55e-05 & \textbf{1.44e-05} & \textbf{1.1229} & 1.2054 & \textbf{1.1229} & 1.2054 & 1.0000 \\
& 128  & 5.52e-05 & \textbf{1.14e-05} & \textbf{0.4541} & 0.5183 & \textbf{0.4542} & 0.5183 & 1.0000 \\
& 256  & 3.62e-05 & \textbf{8.39e-07} & \textbf{0.1733} & 0.1775 & \textbf{0.1733} & 0.1775 & 1.0000 \\
& 512  & 3.76e-05 & \textbf{7.16e-07} & 0.0723 & \textbf{0.0690} & 0.0723 & \textbf{0.0690} & 0.2364 \\
& 1024 & 2.27e-05 & \textbf{2.62e-06} & 0.0328 & \textbf{0.0326} & 0.0328 & \textbf{0.0326} & 0.4459 \\
& 2048 & 1.52e-06 & \textbf{1.13e-09} & \textbf{0.01594} & 0.01648 & \textbf{0.01594} & 0.01648 & 0.9126 \\
& 4096 & \textbf{3.83e-07} & 4.25e-07 & 0.007027 & \textbf{0.006727} & 0.007028 & \textbf{0.006728} & 0.0574 \\
& 8192 & \textbf{3.48e-09} & 3.79e-07 & 0.003379 & \textbf{0.003266} & 0.003379 & \textbf{0.003267} & 0.1613 \\
\bottomrule
\end{tabular}
\caption{Integration results for the Barrier Option across two scenarios. We report FDR-corrected p-values from a one-sided Wilcoxon signed-rank test. P-values below 0.05 and the best methods are \textbf{bolded}.}
\end{table*}

% --- BASKET OPTION RESULTS ---

\begin{table*}[t]
\centering
\small
\label{tab:results_basket}
\begin{tabular}{@{} l l l c c c c c c l @{}}
\toprule
\textbf{Scenario} & \textbf{N} & \multicolumn{2}{c}{\textbf{Squared Bias}} & \multicolumn{2}{c}{\textbf{Variance}} & \multicolumn{2}{c}{\textbf{MSE}} & \textbf{p-value} \\
\cline{3-8}
& & Sobol & LLM & Sobol & LLM & Sobol & LLM & \\
\hline
Low Correlation & 32   & 3.82e-06 & \textbf{5.13e-07} & \textbf{0.1152} & 0.1182     & \textbf{0.1152} & 0.1182     & 0.9968 \\
& 64   & \textbf{6.60e-07} & 1.99e-06 & \textbf{0.0342} & 0.0358     & \textbf{0.0342} & 0.0358     & 1.0000 \\
& 128  & \textbf{1.18e-06} & 1.38e-06 & \textbf{0.0109} & 0.0126     & \textbf{0.0109} & 0.0126     & 1.0000 \\
& 256  & 5.16e-08 & \textbf{1.38e-08} & \textbf{0.0036} & 0.0039     & \textbf{0.0036} & 0.0039     & 1.0000 \\
& 512  & 3.04e-07 & \textbf{1.15e-08} & 0.001259   & \textbf{0.001231} & 0.001259   & \textbf{0.001231} & 0.2381 \\
& 1024 & 2.00e-07 & \textbf{1.79e-08} & 0.000504   & \textbf{0.000473} & 0.000504   & \textbf{0.000473} & \textbf{0.0051} \\
& 2048 & 5.00e-08 & \textbf{6.46e-11} & 0.000229   & \textbf{0.000214} & 0.000229   & \textbf{0.000214} & \textbf{3.79e-04} \\
& 4096 & \textbf{1.12e-10} & 1.32e-09 & 9.886e-05  & \textbf{9.038e-05} & 9.886e-05  & \textbf{9.038e-05} & \textbf{1.68e-04} \\
& 8192 & 1.39e-09 & \textbf{1.72e-10} & 4.740e-05  & \textbf{4.157e-05} & 4.740e-05  & \textbf{4.157e-05} & \textbf{1.92e-11} \\
\cline{1-9}
High Correlation & 32   & \textbf{9.26e-05} & 9.80e-05 & \textbf{0.2483} & 0.2557     & \textbf{0.2484} & 0.2558     & 0.9968 \\
& 64   & \textbf{7.77e-06} & 1.10e-05 & 0.0602     & \textbf{0.0570} & 0.0602     & \textbf{0.0571} & \textbf{2.61e-05} \\
& 128  & 6.94e-06 & \textbf{2.65e-06} & \textbf{0.0163} & 0.0168     & \textbf{0.0163} & 0.0168     & 1.0000 \\
& 256  & \textbf{7.96e-07} & 8.54e-07 & \textbf{0.004907} & 0.004944 & \textbf{0.004908} & 0.004945 & 1.0000 \\
& 512  & \textbf{6.53e-08} & 8.95e-08 & 0.001527   & \textbf{0.001518} & 0.001527   & \textbf{0.001518} & 0.4675 \\
& 1024 & \textbf{3.80e-10} & 1.16e-08 & \textbf{0.000463}   & 0.000447 & \textbf{0.000463}   & 0.000447 & 0.0784 \\
& 2048 & 3.45e-09 & \textbf{1.57e-09} & \textbf{0.0001750}  & 0.0001779 & \textbf{0.0001750}  & 0.0001779 & 0.7419 \\
& 4096 & \textbf{8.68e-10} & 7.39e-09 & 7.132e-05  & \textbf{6.609e-05} & 7.132e-05  & \textbf{6.610e-05} & \textbf{8.20e-04} \\
& 8192 & 6.90e-09 & \textbf{1.17e-10} & 2.598e-05  & \textbf{2.448e-05} & 2.599e-05  & \textbf{2.448e-05} & \textbf{5.32e-03} \\
\cline{1-9}
Mixed Volatility & 32   & 2.19e-04 & \textbf{9.65e-05} & \textbf{0.7753} & 0.7993     & \textbf{0.7755} & 0.7994     & 0.9968 \\
& 64   & \textbf{9.78e-06} & 1.28e-05 & 0.2098     & \textbf{0.2071} & 0.2099     & \textbf{0.2071} & 0.5396 \\
& 128  & 1.95e-05 & \textbf{2.19e-06} & \textbf{0.0647} & 0.0702     & \textbf{0.0647} & 0.0702     & 1.0000 \\
& 256  & 4.42e-06 & \textbf{1.52e-06} & \textbf{0.0211} & 0.0216     & \textbf{0.0211} & 0.0216     & 1.0000 \\
& 512  & 6.51e-07 & \textbf{6.07e-07} & 0.006202   & \textbf{0.006112} & 0.006202   & \textbf{0.006112} & 0.6395 \\
& 1024 & 1.17e-08 & \textbf{5.20e-09} & 0.002089   & \textbf{0.002026} & 0.002089   & \textbf{0.002026} & 0.0784 \\
& 2048 & 1.09e-08 & \textbf{4.65e-11} & 0.000947   & \textbf{0.000932} & 0.000947   & \textbf{0.000932} & \textbf{0.0672} \\
& 4096 & \textbf{3.40e-09} & 2.79e-08 & 0.000447   & \textbf{0.000393} & 0.000447   & \textbf{0.000393} & \textbf{1.45e-10} \\
& 8192 & 1.80e-08 & \textbf{2.19e-12} & 0.000156   & \textbf{0.000143} & 0.000156   & \textbf{0.000143} & \textbf{1.48e-05} \\
\cline{1-9}
Out-of-the-Money & 32   & 2.43e-05 & \textbf{2.29e-09} & \textbf{0.1165} & 0.1205     & \textbf{0.1166} & 0.1205     & 0.9968 \\
& 64   & 2.81e-06 & \textbf{4.90e-09} & \textbf{0.0367} & 0.0387     & \textbf{0.0367} & 0.0387     & 1.0000 \\
& 128  & 2.55e-06 & \textbf{2.01e-07} & \textbf{0.0127} & 0.0143     & \textbf{0.0127} & 0.0143     & 1.0000 \\
& 256  & 5.97e-07 & \textbf{4.26e-10} & \textbf{0.004370} & 0.004627 & \textbf{0.004371} & 0.004627 & 1.0000 \\
& 512  & 2.65e-07 & \textbf{2.67e-08} & 0.001602   & \textbf{0.001553} & 0.001602   & \textbf{0.001553} & \textbf{0.2364} \\
& 1024 & 3.70e-08 & \textbf{2.12e-09} & 0.000652   & \textbf{0.000623} & 0.000652   & \textbf{0.000623} & \textbf{0.0651} \\
& 2048 & 5.68e-09 & \textbf{3.00e-11} & 0.000292   & \textbf{0.000281} & 0.000292   & \textbf{0.000281} & \textbf{0.0856} \\
& 4096 & 1.01e-08 & \textbf{5.67e-09} & \textbf{0.000125}   & \textbf{0.000120} & \textbf{0.000125}   & \textbf{0.000120} & \textbf{0.0027} \\
& 8192 & \textbf{1.58e-11} & 1.66e-09 & 5.949e-05  & \textbf{5.392e-05} & 5.949e-05  & \textbf{5.392e-05} & \textbf{2.53e-06} \\
\bottomrule
\end{tabular}
\caption{Integration results for the 32-dimensional Basket Option across four scenarios. We report FDR-corrected p-values from a one-sided Wilcoxon signed-rank test. P-values below 0.05 and the best methods are \textbf{bolded}.}
\end{table*}

% --- BERMUDAN OPTION RESULTS ---

\begin{table*}[t]
\centering
\label{tab:results_bermudan}
\begin{tabular}{@{} l l l c c c c c c l @{}}
\toprule
\textbf{Scenario} & \textbf{N} & \multicolumn{2}{c}{\textbf{Squared Bias}} & \multicolumn{2}{c}{\textbf{Variance}} & \multicolumn{2}{c}{\textbf{MSE}} & \textbf{p-value} \\
\cline{3-8}
& & Sobol & LLM & Sobol & LLM & Sobol & LLM & \\
\hline
At-the-Money & 32   & \textbf{0.8165} & 0.8272     & \textbf{0.5744} & 0.5826     & \textbf{1.3909} & 1.4098     & 0.9968 \\
& 64   & \textbf{0.2544} & 0.2627     & 0.2403     & \textbf{0.2372} & \textbf{0.4948} & 0.4999     & 1.0000 \\
& 128  & \textbf{0.0698} & 0.0725     & \textbf{0.0941} & 0.0951     & \textbf{0.1639} & 0.1677     & 1.0000 \\
& 256  & \textbf{0.0182} & 0.0192     & \textbf{0.0441} & 0.0445     & \textbf{0.0624} & 0.0637     & 1.0000 \\
& 512  & 0.0044    & \textbf{0.0041}    & \textbf{0.0218} & 0.0221     & \textbf{0.02621} & 0.02622    & 0.4675 \\
& 1024 & 9.87e-04  & \textbf{9.33e-04}  & 0.0118     & \textbf{0.0112} & 0.0128     & \textbf{0.0122}    & \textbf{0.0115} \\
& 2048 & 2.77e-04  & \textbf{2.46e-04}  & 0.005957   & \textbf{0.005713} & 0.006234   & \textbf{0.005960}  & \textbf{0.0018} \\
& 4096 & 5.23e-05  & \textbf{4.25e-05}  & 0.002696   & \textbf{0.002676} & 0.002749   & \textbf{0.002718}  & 0.3863 \\
& 8192 & 1.10e-05  & \textbf{8.45e-06}  & 0.001324   & \textbf{0.001273} & 0.001335   & \textbf{0.001281}  & \textbf{0.0457} \\
\cline{1-9}
In-the-Money & 32   & \textbf{0.8888} & 0.8979     & 0.5629     & \textbf{0.5549} & \textbf{1.4518} & 1.4528     & 0.9968 \\
& 64   & \textbf{0.2226} & 0.2296     & 0.2507     & \textbf{0.2509} & \textbf{0.4733} & 0.4805     & 1.0000 \\
& 128  & \textbf{0.0478} & 0.0489     & 0.1183     & \textbf{0.1169} & 0.1662     & \textbf{0.1658}    & 1.0000 \\
& 256  & \textbf{0.0092} & 0.0093     & 0.0564     & \textbf{0.0563} & \textbf{0.0656} & \textbf{0.0656}    & 1.0000 \\
& 512  & 0.0020    & \textbf{0.0019}    & 0.0284     & \textbf{0.0278} & 0.0304     & \textbf{0.0297}    & 0.4675 \\
& 1024 & \textbf{3.75e-04} & 4.09e-04   & 0.0145     & \textbf{0.0135} & 0.0149     & \textbf{0.0139}    & \textbf{0.0142} \\
& 2048 & \textbf{6.75e-05} & 7.12e-05   & 0.007354   & \textbf{0.006584} & 0.007422   & \textbf{0.006655}  & \textbf{1.22e-05} \\
& 4096 & 1.17e-05  & \textbf{9.36e-06}  & 0.003369   & \textbf{0.003128} & 0.003381   & \textbf{0.003137}  & \textbf{7.39e-04} \\
& 8192 & \textbf{3.56e-06} & 3.86e-06   & 0.001660   & \textbf{0.001523} & 0.001664   & \textbf{0.001527}  & \textbf{1.62e-04} \\
\bottomrule
\end{tabular}
\caption{Integration results for the Bermudan Option across two scenarios. We report FDR-corrected p-values from a one-sided Wilcoxon signed-rank test. P-values below 0.05 and the best methods are \textbf{bolded}.}
\end{table*}